\documentclass[letterpaper, 10pt, conference]{ieeeconf}  

\IEEEoverridecommandlockouts                              

\usepackage{amsmath} 
\usepackage{amssymb}  
\usepackage{cite}
\usepackage{graphicx}
\usepackage[hidelinks]{hyperref}

\usepackage{indentfirst}

\makeatletter
\let\MYcaption\@makecaption
\makeatother
\usepackage[font=footnotesize]{subcaption}
\makeatletter
\let\@makecaption\MYcaption
\makeatother

\usepackage{tikz}
\usepackage{pgfplots}
\usetikzlibrary{external}
\usetikzlibrary{matrix, decorations.pathreplacing, calc, positioning, fit, fadings, shapes.arrows, arrows.meta, shapes.geometric, backgrounds}
\usetikzlibrary{shadings}
\usepackage{tikzscale}

\usepackage{algorithm}
\usepackage{algorithmicx}
\usepackage[noend]{algpseudocode}
\usepackage[T1]{fontenc}
\usepackage{inconsolata}
\usepackage{comment}
\usepackage{multirow}
\usepackage{hyperref}

\newcommand\copyrighttext{\footnotesize \textcopyright~2024 IEEE. Personal use of this material is permitted. Permission from IEEE must be obtained for all other uses, in any current or future media, including reprinting/republishing this material for advertising or promotional purposes, creating new collective works, for resale or redistribution to servers or lists, or reuse of any copyrighted component of this work in other works.%
DOI: \href{https://ieeexplore.ieee.org/document/10588674}{10.1109/IV55156.2024.10588674}
}

\newcommand\copyrightnotice{%
    \begin{tikzpicture}[remember picture,overlay]%
     \node[%
        anchor=south, %
        yshift=10pt%
    ] at (current page.south)%
     {\fbox{\parbox{\dimexpr\textwidth-\fboxsep-\fboxrule\relax}{\copyrighttext}}};%
     \end{tikzpicture}%
}

\title{\LARGE \bf
A Graph Neural Network Approach for Solving the Ranked Assignment Problem in Multi-Object Tracking
}

\author{Robin Dehler$^{1}$, Martin Herrmann$^{1}$, Jan Strohbeck$^{1}$ and Michael Buchholz$^{1}$
	\thanks{Parts of this work have been financially supported by the Federal Ministry of Education and Research (project AUTOtech.\emph{agil}, FKZ 01IS22088W). 
 Parts of this research have been conducted as part of the PoDIUM project and other parts as part of the EVENTS project, which both are funded by the European Union under grant agreement No. 101069547 and No. 101069614 respectively. Views and opinions expressed are however those of the authors only and do not necessarily reflect those of the European Union or European Commission. Neither the European Union nor the granting authority can be held responsible for them.}
	\thanks{$^{1}$The authors are with the Institute of Measurement, Control and Microtechnology, Ulm University, D-89081 Ulm, Germany.\newline
    {E-mail addresses: \{firstname\}.\{lastname\}@uni-ulm.de}%
	}%
}
\begin{document}

\maketitle
\thispagestyle{empty}
\pagestyle{empty}

\begin{abstract}
Associating measurements with tracks is a crucial step in Multi-Object Tracking (MOT) to guarantee the safety of autonomous vehicles.
To manage the exponentially growing number of track hypotheses, truncation becomes necessary.
In the $\delta$-Generalized Labeled Multi-Bernoulli ($\delta$-GLMB) filter application, this truncation typically involves the ranked assignment problem, solved by Murty's algorithm or the Gibbs sampling approach, both with limitations in terms of complexity or accuracy, respectively.
With the motivation to improve these limitations, this paper addresses the ranked assignment problem arising from data association tasks with an approach that employs Graph Neural Networks (GNNs).
The proposed Ranked Assignment Prediction Graph Neural Network (RAPNet) uses bipartite graphs to model the problem, harnessing the computational capabilities of deep learning.
The conclusive evaluation compares the RAPNet with Murty's algorithm and the Gibbs sampler, showing accuracy improvements compared to the Gibbs sampler.
\end{abstract}
\copyrightnotice
\section{Introduction}
Reliable multi-object tracking (MOT) is an important component for many technical applications, e.g., for robotics or autonomous driving.
As part of environmental awareness modeling, MOT is faced with the task of tracking multiple objects over time given uncertainty in the detections \cite{bar_11, mahler_07}.
Different classical approaches to handle the tracking problem can be summarized into the categories Joint Probabilistic Data Association (JPDA), Multi-Hypothesis Tracking (MHT) \cite{bar_11} and approaches using Random Finite Sets (RFS) and the Finite Set Statistics (FISST) framework \cite{mahler_07}.

Within the RFS approach, a Bayes optimal MOT filter was introduced using Generalized Labeled Multi-Bernoulli (GLMB) RFSs \cite{vo_13}.
For a real-time capable implementation, the GLMB filter was extended to the $\delta$-Generalized Labeled Multi-Bernoulli ($\delta$-GLMB) filter \cite{vo_14}.
In the update step of the $\delta$-GLMB filter, it is necessary to associate detections with objects or tracks \cite{vo_14}.
When organizing both the detections and tracks in a cost matrix, the association task results in a ranked assignment problem that is solved optimally with Murty's algorithm \cite{murty_68}.
Solving the ranked assignment problem poses high computational complexity.
The association task is NP-hard when dealing with Multi-Sensor MOT (MS-MOT), where the association involves multiple tracks, measurements, and sensors \cite{vo_19}.
Multi-sensor setups are crucial for creating a comprehensive environment model in autonomous driving, hence the importance of MS-MOT~\cite{buchholz_22}.

The $\delta$-GLMB filter offers a mathematically optimal solution for MOT \cite{vo_14}.
Because of the filter's complexity, however, the literature includes various mathematically suboptimal approaches, where the entire tracking task is executed using Deep Learning (DL) models, e.g., SORT and StrongSORT \cite{wojke_17, du_18}, MOTS \cite{voigtlaender_19} or SMILEtrack \cite{wang_22}.

Motivated by the promising results of DL approaches for tracking applications, this work aims to combine the advantages of the $\delta$-GLMB filter with a GNN model that solves the computationally expensive data association task.
Although being especially challenging for MS-MOT, this paper focuses on solving two-dimensional assignment problems, to form the basic theory for an entire class of DL methods for solving ranked assignment problems in single- and multi-sensor applications. Moreover, we propose a suitable DL model based on Graph Neural Networks (GNNs).
For that, we represent the ranked assignment problem using bipartite graphs, where the source and target nodes align with the rows and columns of the cost matrix, respectively.
The objective is then framed as a weighted bipartite matching problem \cite{burkhard_12}.

Our main contributions are as follows:
\begin{itemize}
    \item An introduction to the fundamental theory of describing assignment problems in MOT that can be used to solve the problem with a new class of DL-based algorithms.
    \item A specific GNN framework to predict ranked assignments denoted Ranked Assignment Prediction Graph Neural Network (RAPNet).
    \item A new score called \textit{weighted position} (wp) for evaluating the performance of approximation algorithms for the ranked assignment problem focusing on the position of the assignments.
\end{itemize}

\section{Background} \label{sec:SOTA}
For the considered MOT use case, the notion of RFSs is used to model the state of multiple objects and multiple measurements.

\subsection{Multi-Object Generalized Labeled Multi-Bernoulli Filter}
Taking the common notations from \cite{vo_13}, a labeled state of an object is given by $\mathbf{x}=(x, l)\in\mathbb{X}\times\mathbb{L}$ with the state space $\mathbb{X}$ and label space $\mathbb{L}$ containing state vectors $x$ and labels $l$, respectively.
Then, the labeled states of all objects can be combined to the labeled multi-object state $\mathbf{X} = \{\mathbf{x}^{(1)}, \ldots, \mathbf{x}^{(n)}\}$.
Similarly, $m$ measurements $z$ form the multi-measurement state $Z=\{z^{(1)}, \ldots z^{(m)}\}$.
With $\mathcal{L}(\mathbf{X})$ as the set of labels of $\mathbf{X}$ and the distinct label operator $\Delta(\mathbf{X})$ that removes infeasible multi-object states and label sets, the multi-object probability density function of a GLMB RFS is defined by
\begin{align} \label{glmb_eq}
    \boldsymbol{\pi}(\mathbf{X}) = \Delta(\mathbf{X})\sum_{c\in\mathbb{C}}w^{(c)}(\mathcal{L}(\mathbf{X}))\left[p^{(c)}\right]^{\mathbf{X}},
\end{align}
where $c\in\mathbb{C}$ systematically lists the GLMB hypotheses consisting of the product of single-object densities $p^{(c)}(\mathbf{x})$ with corresponding weights $w^{(c)}(\mathcal{L}(\mathbf{X}))$.
In (\ref{glmb_eq}), multiple copies of identical spatial distributions are considered.
For a $\delta$-GLMB, the probability density function in (\ref{glmb_eq}) is adjusted so that these duplicate copies are removed \cite{vo_13}.\footnote{Since the following considerations are similar for both GLMB and $\delta$-GLMB RFSs, the detailed replacements and resulting equations for the latter are excluded.}

Given a GLMB or $\delta$-GLMB filtering density $\boldsymbol{\pi}_k(\mathbf{X}_k)$ at time step $k$, the GLMB prediction and update densities $\boldsymbol{\pi}_{k+1|k}(\mathbf{\mathbf{X}}_{k+1}|Z_{1:k})$ and $\boldsymbol{\pi}_{k+1}(\mathbf{X}_{k+1}|Z_{1:k+1})$ can be calculated using the Chapman-Kolmogorov equation and Bayes rule.
In the update step, each hypothesis creates a new set of hypotheses and the number of hypotheses grows intractably high.
Thus, for a real-time capable implementation, it is necessary to truncate the number of resulting hypotheses at each time step.
Using the ranked assignment problem, the weights corresponding to the hypotheses can be exploited to calculate the highest weighted hypotheses only, without the need to compute all possible new ones.

\subsection{Ranked Assignment Problem}
Every possible association of a set of measurements $Z=\{z_1,\ldots,z_{|Z|}\}$ with a set of tracks $I=\{l_1,\ldots,l_{|I|}\}$ has an individual cost value $c_{ij}$ that depends on the multi-target model, e.g., Gaussian Mixture or sequential Monte Carlo approximation \cite{vo_14}.
Furthermore, it is possible that a track $l_i$ is misdetected with a corresponding cost value of $c_i$.
All the cost values can be organized in a cost matrix $C_Z$ with row $i$ corresponding to track $i\in\{1,\ldots,|I|\}$ and column $j$ to either a measurement for $1\leq j \leq |Z|$ or a misdetection for $|Z|+1 \leq j \leq |Z|+|I|$:
\vspace*{-.3em}
\begin{align}\label{eq:cz}
C_Z& = \begin{tikzpicture}[baseline=-0.5ex]
    \matrix [matrix of math nodes,left delimiter={[},right delimiter={]},inner sep=0.5pt, row sep={0.05cm}, column sep=0.05cm, font=\normalsize, nodes={scale=.99}, ampersand replacement=\&] (m) {
        c_{11} \& \& \dots \& \& c_{1|Z|} \& \, \& c_1 \& \infty \& \dots  \& \& \infty \& \text{} \\
        \& \ddots \& \& \& \& \, \& \infty \& c_2 \& \ddots \& \& \& \\
        \vdots \&  \& c_{ij} \& \& \vdots \& \, \& \vdots \& \ddots \& \ddots \& \& \vdots \& \\
        \& \& \& \ddots \& \& \,\& \& \& \& c_{|I|-1} \& \infty \& \\
        c_{|I|1} \& \& \dots \& \& c_{|I||Z|} \& \, \& \infty \& \& \dots  \& \infty \& c_{|I|} \& \\
    };

    \node[
    fit={($(m-1-1)+(-0.5em,0)$)($(m-1-6.west)+(-0.5em,0)$)},
    inner xsep=0,
    above delimiter=\{,
    label={[yshift=0.3cm]\footnotesize{detected}}
    ] {};
    \node[
    fit={($(m-1-7)+(-0.5em,0)$)($(m-1-12.east)+(-0.5em,0)$)},
    inner xsep=0,
    above delimiter=\{,
    label={[yshift=0.3cm]\footnotesize{misdetected}}
    ] {};
\end{tikzpicture}
\end{align}
The entry $c_{ij}$ in the \textit{detected} part of (\ref{eq:cz}) is the cost of associating track $i$ with measurement $j$ and the entry $c_i$ in the \textit{misdetected} part the cost value for misdetecting track $i$.
The $\infty$-values of the \textit{misdetected} part of the matrix imply invalid associations, i.e., only one cost value indicates the misdetection of track $i$.
Similarly, the \textit{detected} part of $C_Z$ can also contain $\infty$-values that are created through a gating step to erase unlikely associations.

An assignment matrix $S$ of the same shape $|I|\times(|Z|+|I|)$, with entries $s_{ij}$ of either $0$ or $1$, models the association of tracks with measurements subject to
\begin{subequations}\label{ap_conditions}
\begin{align}
    &\textstyle\sum_{i=1}^{|I|} s_{ij} \leq 1, & (j=1, ..., |Z|+|I|) \\ 
    &\textstyle\sum_{j=1}^{|Z|+|I|} s_{ij} = 1, & (i=1, ..., |I|)
\end{align}
\end{subequations}
The conditions (\ref{ap_conditions}) define characteristics for $S$ where each row has the sum $1$ and each column the sum $0$ or $1$, ensuring that exactly one measurement or misdetection is associated with one track\cite{vo_14}.
Given a cost matrix $C_Z$ and an assignment matrix $S$, the overall cost of an assignment is given by the Frobenius inner product
\begin{equation}\label{frobenius}
    Tr(S^TC_Z) = \sum_{i=1}^{|I|}\sum_{j=1}^{|Z|+|I|}c_{ij}s_{ij}.
\end{equation}
Note that since $j\in\{1,\ldots,|Z|+|I|\}$, the cost value $c_{ij}$ in (\ref{frobenius}) can also correspond to an entry $c_i$ in (\ref{eq:cz}).
The optimal assignment is the one that minimizes (\ref{frobenius}), which can be calculated, e.g., with the Hungarian method \cite{kuhn_55} or the Jonker-Volgenant algorithm \cite{jonker_87}.
For MOT, it is beneficial to calculate more than just the best assignment, which results in the ranked assignment problem to find a set of $k$ minimal cost assignments with increasing order \cite{murty_68}.

Equivalent to representing the costs in a cost matrix, the association of measurements with tracks can be modeled by a bipartite graph
\begin{align}
    \mathcal{G} = \{\mathcal{V}_s, \mathcal{V}_t, \mathcal{E}\},
\end{align}
where the tracks are mapped to the source nodes $v_s\in\mathcal{V}_s$ and the measurements and misdetections to the target nodes $v_t\in\mathcal{V}_t$.
The cost values $c_{ij}$ and $c_{i}$ for association of tracks with measurements and misdetections, respectively, are modeled by the edge attributes $a_{ij}$ of the edges $e_{ij}\in\mathcal{E}$ that connect source nodes $v_s^{(i)}$ with target nodes $v_t^{(j)}$. The costs $c_i$ are matched with edges $e_{i,i+|Z|}$ .
The $\infty$-values are modeled by excluding edges for the corresponding pairs of $v_s$ and $v_t$.
Similar to the assignment matrix $S$, an association map can be represented by classifying the edges as either $0$ or $1$ with the following constraints:
(i) From the set of edges originating from a source node, exactly one edge is classified as $1$ and the rest as $0$.
(ii) From the set of edges connected to a target node, at most one edge is classified as $1$.
When represented as a bipartite graph, the ranked assignment problem is a weighted bipartite matching problem \cite{burkhard_12}.

In \cite{vo_14}, the ranked assignment problem for the truncation of $\delta$-GLMB hypotheses is solved using Murty's algorithm \cite{murty_68}.
A more efficient algorithm is proposed in \cite{vo_17}, where the optimal solution of the ranked assignment problem is approximated using a method based on Gibbs sampling.
The Gibbs sampling approach, however, has the drawback that it is less accurate.
The architecture proposed in this paper also approximates the optimal solution of the ranked assignment problem by using the bipartite graph formulation of the problem in order to use GNNs to predict the assignments.

\subsection{Graph Neural Networks}
GNNs are a powerful tool to solve tasks on data that is represented with graphs \cite{scarselli_09}, e.g., graph or node classification and link prediction.
Using the notion of message passing, GNN layers aggregate information about the neighborhood of nodes using an activation function that combines the node feature information with the edge features \cite{ma_21}.
The Graph Attention Network (GAT), e.g., updates each node feature similar to the attention mechanism \cite{velickovic_18}, where an individual importance score is calculated for each neighboring node.

\subsection{Related Work}
In literature, there exist some approaches for solving assignment problems with deep learning methods.
In \cite{lee_18}, a method is proposed where the assignment problem is decomposed into sub-assignment problems that are independently solved with deep neural networks.
In \cite{aironi_22}, a GNN architecture is developed for solving the assignment problem.
However, for the GNN framework, quadratic cost matrices are required, which limits the flexibility of the matrices that can be used.
In \cite{liu_22}, Liu \textit{et al.} introduce GLAN, which is another approach based on GNNs that is able to handle graphs of different sizes.
The GLAN network is also evaluated in an MOT scenario, where the network combined with the Faster R-CNN object detector achieves high MOTA and MOTP scores.
Nonetheless, all of the mentioned systems are only trained to obtain the best assignment, not to solve the ranked assignment problem.
Solving the ranked assignment problem with DL methods has not yet been researched.

\section{Ranked Assignment Prediction} \label{sec:RAPNet}
The proposed framework to predict the ranked assignments consists of a graph creation module, the Ranked Assignment Prediction Graph Neural Network (RAPNet) and a post-processing stage to extract assignments from the RAPNet output.
The three modules are explained in the following.

\subsection{Graph Creation}
Like the known methods, our approach uses the cost matrix representation as input data.
The first step to use a GNN for the prediction is to transform the cost matrices to bipartite graphs.
The transformation includes source and target node creation with corresponding node features $x_s$ and $x_t$, respectively, and the creation of edge indices $e_{ij}$ that indicate the connection of source nodes $i$ with target nodes $j$ with edge attributes $a_{ij}$.
The edge attributes can directly be taken from the values of the cost matrix.
However, because of the dynamic number of tracks and measurements, the cost values of the rows and columns can not directly be taken for the node features, since the dimensions of the features need to be consistent.
Thus, for each row and column, the following five features were identified to provide a good representation of the input data:
(i) The ratio of non-$\infty$ values to the length of the row or column, respectively, and the values (ii) \textit{min}, (iii) \textit{max}, (iv) \textit{mean} and (v) \textit{l2-norm} are aggregated without consideration of the $\infty$ entries.
The number of source nodes $\nu_s$ and target nodes $\nu_t$ correlate with the number of rows and columns of the input matrices, respectively.
In the considered use case, the input matrices are of the form of $C_Z$ from Eq.~(\ref{eq:cz}). From this, $\nu_s=|I|$ and $\nu_t = |I|+|Z|$ follow. 

\subsection{Ranked Assignment Prediction Graph Neural Network}
\begin{figure*}[t]
    \centering
    \input{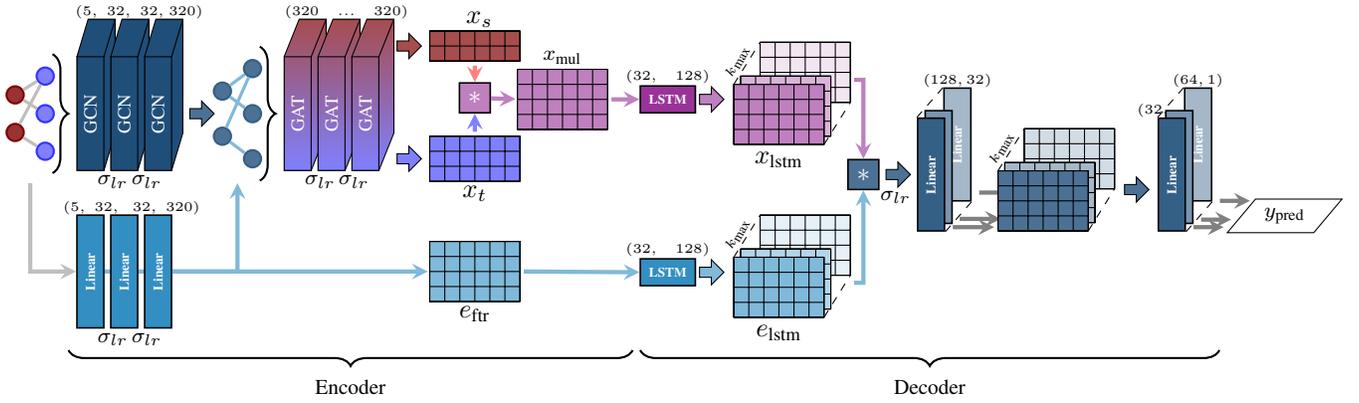}
    \vspace{-15pt}
    \caption{Architecture of the RAPNet. Left part: Encoding of the node and edge features. Before the multiplication step of $x_s$ and $x_t$, the feature vectors are indexed with the respective edge index vectors to fit the dimensions. Right part: Decoding using LSTM layers and creation of predicted assignments. The channel sizes $C_{\text{enc}}=32$, $C_{\text{lstm}}=128$ and $C_{\text{dec}}=32$ are shown above each neural network layer.  The final output $y_{\text{pred}}$ is of shape $|\mathcal{E}|\times k_{\text{max}}$.}
    \label{architecture}
    \vspace{-0.4cm}
\end{figure*}

The RAPNet takes the created bipartite graphs as inputs.
The task of predicting assignments is modeled as a binary classification task on the graph edges.
The most effective network architecture, as determined through comprehensive research, is depicted in Fig.~\ref{architecture}, following an encoder-decoder structure.
The encoder processes both the node and edge features through GNN and Linear layers, respectively.
The encoded features are then combined through element-wise multiplication and split into multiple assignment predictions in the decoder.
For a better training, the inputs, i.e., the node and edge features, are normalized to the range $[0, 1]$.

\subsubsection*{Encoder}
As shown in Fig.~\ref{architecture}, the node and edge features of the input graphs are first updated using Graph Convolutional Network (GCN) layers \cite{kipf_16} and Fully Connected (Linear) layers, respectively, with leaky ReLU activation functions ($\sigma_{lr}$) in between.
After the GCN layers, the updated features are combined and fed through Graph Attention Network (GAT) layers \cite{velickovic_18}, again using leaky ReLU activations.
The last GAT layers output encoded features $x_s$ and $x_t$.
Through indexing with the corresponding edge indices, $x_s$ and $x_t$ are multiplied element-wise to combine the features to $x_{\text{mul}}$.
Both GCN and GAT layers only update the target node features of bipartite graphs.
For updating the source nodes, the source and target nodes are mirrored for each layer.
Thus, each of the shown GCN and GAT layer blocks consist of two layers to update both $x_s$ and $x_t$, respectively.

Dependent on the MOT application, a different number of assignments can potentially be required for each cost matrix and consequently for each graph.
To be able to use batched graphs and also to better train the network, a fixed number of output assignments $k_{\text{max}}=10$ was chosen.
This has the disadvantage that it is not possible to optimize the network for tasks with $k>k_{\text{max}}$, but has the advantages that each of the $k_{\text{max}}$ solutions are better optimized for predicting one particular solution and that for $k\leq k_{\text{max}}$, the network predicts more than the necessary assignments which increases the possibility of predicting the optimal ones.
Thus, the GAT layers and the last GCN and Linear layer in the encoder have an output dimension that is the product of the number of channels for the encoder layers $C_{\text{enc}}$ and $k_{\text{max}}$.

\subsubsection*{Decoder}
The decoder architecture is depicted in the right part of Fig.~\ref{architecture}.
After reshaping, the encoded node features $x_{\text{mul}}$ and edge features $e_{\text{ftr}}$ are fed into separate Long Short-Term Memory (LSTM) Recurrent Neural Networks (RNN) to create $k_{\text{max}}$ different features.
Then, the LSTM outputs $x_{\text{lstm}}$ and $e_{\text{lstm}}$ are multiplied element-wise.
After a leaky ReLU activation, two separate groups of linear layers are applied to each of the $k_{\text{max}}$ solutions.
In between the two groups, the output of the first group for layers $i\in[2, k_{\text{max}}]$ is concatenated with the output of layer $i-1$ before the second group to create an additional dependency between the assignments. Thus, the second up to the last linear layer in the second group receive two times the number of input features compared to the first layer, as indicated in Fig. \ref{architecture}.
The resulting output $y_{\text{pred}}$ of the RAPNet is of shape $|\mathcal{E}|\times k_{\text{max}}$.
Each of the $k_{\text{max}}$ columns of $y_{\text{pred}}$ represents an individual predicted assignment matrix.

\subsubsection*{Loss function}
Since the problem is a binary classification task, a Sigmoid activation is employed on the output of the neural network $y_{\text{pred}}$.
The targets of the neural network $y$ represent the optimal assignments calculated with Murty's algorithm.
Then, a (weighted) Binary Cross-Entropy (BCE) function calculates the loss between the network output $y_{\text{pred}}$ and the optimal solutions $y$.
The weights for the two classes $0$ and $1$ are determined based on their ratio to account for class imbalance due to sparsity.

\subsection{Post-processing Module}
The output $y_{\text{pred}}$ of the RAPNet can directly be converted to $k_{\text{max}}$ predicted assignment matrices $A_{\text{pred}}^{(i)}$, $i\in[1,k_{\text{max}}]$.
From each assignment matrix $A_{\text{pred}}^{(i)}$, the predicted assignment $a_{\text{pred}}^{(i)}$ is extracted by taking the maximum value of each row.
As a consequence of inaccurate predictions, it is possible that invalid assignments are created, i.e., that a column is assigned twice.
Thus, a greedy strategy expands the number of chosen assignments.
The greedy strategy exploits the imperfect output $A_{\text{pred}}^{(i)}$ by taking more than only the maximum value of each row.
It additionally uses other entries where the predicted values are greater than a threshold $\theta_{\text{sig}}$ to increase the likelihood of getting valid assignments.
Since the sigmoid activation function creates values between $0$ and $1$, we set $\theta_{\text{sig}} = 0.5$.

Algorithm \ref{alg:gpp} summarizes the post-processing procedure.
The maximum value of each row is taken and set to $a_{\text{found}}$ (code line \ref{argmax}).
Then, the number of entries that are greater than $\theta_{\text{sig}}$ is determined for each row and stored in $a_{\text{sum}}$ (line 4).
As long as there are rows with multiple assignments, which is equivalent to $a_{\text{sum}}$ containing entries $\geq2$, the one with the most entries $\geq\theta_{\text{sig}}$ is taken and the highest entry is set to $0$.
Then, a new possible assignment $a_{\text{add}}$ is taken from the maximum entries of the modified predicted assignment matrix and added to $a_{\text{found}}^{(i)}$.
This is repeated until all redundant predictions in the rows of $A_{\text{pred}}^{(i)}$ are added to $a_{\text{found}}$ (lines 5-9).
Finally, the invalid solutions in $a_{\text{found}}$ are removed (line 10) and all valid solutions from each $A_{\text{pred}}^{(i)}$ are concatenated and returned (lines 11-13).
The steps in the for-loop can be done independently for all $A_{\text{pred}}^{(i)}$.
Thus, the for-loop is parallelized to reduce the computational overhead.

\begin{algorithm}[t]
    \caption{Greedy Post Processing}
    \label{alg:gpp}
    \hspace*{\algorithmicindent} \textbf{Input:} $y_{\text{pred}}$, $k_{\text{max}}$, \textbf{Output:} Assignments $a_{\text{valid}}$
    \begin{algorithmic}[1] \label{postpr}
        \State Get $A_{\text{pred}}^{(i)}$, $i\in[1,k_{\text{max}}]$ from $y_{\text{pred}}$
        \For{Each $A_{\text{pred}}^{(i)}$}
        \State $a_{\text{pred}}^{(i)}= \mathrm{argmax}_{\text{rows}} A_{\text{pred}}^{(i)}$, $a_{\text{found}}\gets \{a_{\text{pred}}^{(i)}\}$ \label{argmax}
        \State $A_{\text{round}}\gets$ Round $A_{\text{pred}}^{(i)}$, $a_{\text{sum}}\gets$ $\sum_{\text{rows}} A_{\text{round}}$
        \While{Entries $\geq 2$ in $a_{\text{sum}}$}
            \State Row $j\gets \mathrm{argmax}(\sum_{\text{rows}}A_{\text{round}})$
            \State Set max value of row $j$ in $A_{\text{pred}}^{(i)}$ to $0$
            \State $a_{\text{add}}=\mathrm{argmax}_{\text{rows}}A_{\text{pred}}^{(i)}$, $a_{\text{found}} =a_{\text{found}}\cup \{a_{\text{add}}\}$
            \State Calculate new $A_{\text{round}}, a_{\text{sum}}$ from updated $A_{\text{pred}}^{(i)}$
        \EndWhile
        \State $a_{\text{valid}}^{(i)}\leftarrow$ Valid solutions of $a_{\text{found}}$
        \EndFor
        \State $a_{\text{valid}} \gets$ Concatenate all $a_{\text{valid}}^{(i)}$
        \State Keep at most $k_{\text{max}}$ solutions with lowest cost
        \State \textbf{Return} $a_{\text{valid}}$
    \end{algorithmic}
\end{algorithm}

\section{Experiments} \label{sec:eval}
This section summarizes the parameters of the RAPNet and compares its performance to the optimal solution of Murty's algorithm and the solutions calculated with the Gibbs sampling method.

\subsection{Training Setup}
The RAPNet is initialized with $32$ encoder channels, $128$ LSTM channels and $32$ decoder channels (see also Fig. \ref{architecture}).
The training involved $20$ epochs with an AdamW optimizer \cite{loshchilov_18} with weight decay $\lambda=0.001$ and the cosine annealing learning rate scheduler \cite{loshchilov_16} with initial learning rate of $\gamma = 0.001$ and minimum value of $\gamma_{\text{min}} = 0.0001$.
The data used for the training included both synthetic graphs as well as graphs that were extracted from a simulated MOT scenario that was introduced in \cite{herrmann_21} for the evaluation of the PM-$\delta$-GLMB filter.
Since the primary focus is solving the ranked assignment problem for the $\delta$-GLMB update step, the synthetically created graphs are designed to be similar to the ones resulting from the cost matrices of the MOT simulation.
Thus, the matrices and equivalently the graphs are also designed to have a detected and misdetected part.
The entries were created using a Gaussian Mixture model with two components with mean $[-2.5, 0.5]$, covariance $[0.5, 1.5]$, and equal weights, respectively.
These values were chosen to be alike to the values of the simulation data.
Lastly, possible $\infty$-values of the detected part were also considered by randomly adding $\infty$-values with a probability threshold $\vartheta\in[0,1]$. A threshold of $0$ means that none of the values are $\infty$, a threshold of $1$ means that all values are set to $\infty$.

Several parameters of the synthetic graphs can be modified, i.e., the threshold $\vartheta$, the value $k$, and the row number $\nu_s$.
The column number $\nu_t$ of the detected part of the cost matrix is chosen to be in the range $[\min(1, \nu_s-1), \nu_s+4]$.
For the training dataset, the value of $k$ is sampled from a Poisson distribution with mean $4$.
Note that the mean value of $k$ of the simulated data is $3.5692$.
The threshold $\vartheta$ is increased from $0.1$ to $0.9$ in steps of $0.1$ and $\nu_s$ from $1$ to $15$.

\subsection{Evaluation Setup}
The first part of the evaluation compares the performance of the RAPNet alone (RAPNet-a) and the RAPNet with the post-processing stage (RAPNet-PP) to the Gibbs sampler and Murty's algorithm using two different parameter sweeps on synthetically created data with changing parameters $\nu_s$ and the maximum number of $k$.
Second, the performance on a validation set including only the simulated data is compared.
Lastly, the needed computational times of the three methods to create assignments are compared.

Three different scores are used for the comparison.
First, the accuracy of each single assignment w.r.t.\ the optimal assignments is calculated by comparing each of the $k$ optimal assignments to the predicted ones.
Second, the overall costs of the chosen assignments are used.
If less than $k$ assignments were found, the missing assignments are penalized with $c_{\text{max}} + 0.1$, with $c_{\text{max}}$ being the cost of the worst assignment.
The first assignments are particularly important for the MOT application because they contain the most likely associations.
Thus, a new score called \textit{weighted position} (\textit{wp}) is proposed, which evaluates the position of the predicted assignments compared to the optimal assignments:
\begin{align}\label{eq:wp}
    wp = \sum_{i=1}^{k}w_i\kappa_i
\end{align}
In (\ref{eq:wp}), $\kappa_i$ rates the position of the predicted assignment $a_{\text{pred}}^{(i)}$ to the optimal one $a^{(i)}$ with
\begin{align*}
\kappa_i =
    \begin{cases}
        3, \text{\ if\ } a_{\text{pred}}^{(i)}=a^{(i)} \\
        2, \text{\ if\ } a_{\text{pred}}^{(i)}\in\left\{a^{(i-\rho)},\ldots,a^{(i-1)},a^{(i+1)},\ldots,a^{(i+\rho)}\right\} \\
        1, \text{\ if\ } a_{\text{pred}}^{(i)}\in \left\{a^{(j)}\setminus \{a^{(i-\rho)}, \ldots, a^{(i+\rho)}\}, j\in[1,k]\right\}\\
        0, \text{\ otherwise},
    \end{cases}
\end{align*}
where $\rho$ is adjustable and in our case set to $2$. The position is weighted with the weight value
\begin{equation}
    w_i = \frac{2\cdot(k+1-i)}{k\cdot(k+1)}
\end{equation}
to emphasize the first assignments.
The score is able to better validate whether generally good assignments were found compared to the accuracy, even if not all optimal ones are generated.
The highest value for $wp$ is $3$ that results from the assignments of Murty's algorithm.

\subsection{Performance Evaluation}
\begin{figure*}[t]
    \vspace{3pt}
    \begin{subfigure}{.48\textwidth}
\begin{tikzpicture}

\definecolor{darkgray176}{RGB}{176,176,176}
\definecolor{custom-green}{RGB}{0,158,115}
\definecolor{custom-darkblue}{RGB}{0,114,178}

\begin{axis}[
tick pos=left,
tick label style={black, font=\tiny},
x grid style={darkgray176},
xmajorgrids,
xmin=1.1, xmax=20.9,
xtick style={color=black},
xtick distance=2,
y grid style={darkgray176},
ylabel={\scriptsize Accuracy},
ylabel style={yshift=-16pt},
ymajorgrids,
ymin=-0.0390233128834356, ymax=1.0494773006135,
ytick style={color=black},
width=1\textwidth,
height=0.56\textwidth,
legend columns = 1,
legend style={font=\tiny, at={(-0.1,0.95)}, line width=0.2pt,
        },
legend image post style={scale=.8},
]
\addplot [semithick, custom-green, mark=o, mark size=1.5, mark options={solid,fill opacity=0}]
table {%
2 0.942814814814815
3 0.944486772486772
4 0.942550264550265
5 0.942825396825397
6 0.943291005291005
7 0.943132275132275
8 0.944814814814815
9 0.942952380952381
10 0.942507936507937
11 0.944095238095238
12 0.943830687830688
13 0.942634920634921
14 0.943883597883598
15 0.942751322751323
16 0.943439153439153
17 0.942793650793651
18 0.942645502645503
19 0.943132275132275
20 0.942571428571429
};
\addplot [semithick, dashed, custom-green, mark=square, mark size=1.5, mark options={solid,fill opacity=0}]
table {%
2 0.786882811533226
3 0.791033867651067
4 0.78887179996031
5 0.789745559051607
6 0.788835733584513
7 0.788709697412949
8 0.786708782360956
9 0.790309188921247
10 0.788970633693972
11 0.789779829457556
12 0.787285487550838
13 0.788327407425761
14 0.790444683988155
15 0.789443040483623
16 0.789567609770606
17 0.786879353511316
18 0.786580456566707
19 0.787054013875124
20 0.78674358720412
};
\addplot [semithick, dash dot, custom-green, mark=diamond, mark size=1.5, mark options={solid,fill opacity=0}]
table {%
2 nan
3 0.536353584556077
4 0.533821783065582
5 0.533828862220263
6 0.534047037320253
7 0.532931763013575
8 0.532112019895966
9 0.536999471557881
10 0.534155741429509
11 0.533073024840076
12 0.533773874066872
13 0.534211041503524
14 0.537393961062177
15 0.537089262404719
16 0.534487465673128
17 0.531618759455371
18 0.530252675433714
19 0.534752512801062
20 0.530981012658228
};
\addplot [semithick, dotted, custom-green, mark=triangle, mark size=1.5, mark options={solid,fill opacity=0}]
table {%
2 nan
3 nan
4 0.330846625766871
5 0.332369467354517
6 0.335161553081727
7 0.333963051105277
8 0.334057937851815
9 0.336614943278133
10 0.333745520088865
11 0.332484725050916
12 0.329192173808666
13 0.333457665050354
14 0.339802064571108
15 0.335698617072008
16 0.336090339149483
17 0.331612903225806
18 0.334662236987818
19 0.33613013384251
20 0.331008730986766
};
\addplot [semithick, custom-darkblue, mark=o, mark size=1.5, mark options={solid,fill opacity=0}]
table {%
2 1
3 1
4 1
5 1
6 1
7 1
8 1
9 1
10 1
11 1
12 1
13 1
14 1
15 1
16 1
17 1
18 1
19 1
20 1
};
\addplot [semithick, dashed, custom-darkblue, mark=square, mark size=1.5, mark options={solid,fill opacity=0}]
table {%
2 0.126568493998268
3 0.152899933107053
4 0.16662532248462
5 0.175767410747467
6 0.184802815854052
7 0.189411312615438
8 0.192877492877493
9 0.198004833013198
10 0.199307573415765
11 0.200183166049925
12 0.201616903085011
13 0.206097696948674
14 0.203434561201353
15 0.206010604033497
16 0.206664931652849
17 0.207109392429538
18 0.205962865247173
19 0.208746283448959
20 0.209121521243934
};
\addplot [semithick, dash dot, custom-darkblue, mark=diamond, mark size=1.5, mark options={solid,fill opacity=0}]
table {%
2 nan
3 0.028957355885605
4 0.0375523739105106
5 0.0470618876845712
6 0.0526512113695283
7 0.0569990110581678
8 0.0606686866210134
9 0.0657143344186284
10 0.0686312458977768
11 0.0690371354988785
12 0.0722937231185011
13 0.0735937092143044
14 0.0744954009165665
15 0.0769503488952992
16 0.0762979973983844
17 0.0787483079863046
18 0.0788875284239986
19 0.0813736645805677
20 0.0793670886075949
};
\addplot [semithick, dotted, custom-darkblue, mark=triangle, mark size=1.5, mark options={solid,fill opacity=0}]
table {%
2 nan
3 nan
4 0.00991411042944785
5 0.0163309565274538
6 0.0206081998370893
7 0.02453445864717
8 0.0276188732781108
9 0.0292302342765245
10 0.0328002850376208
11 0.0337067209775967
12 0.0351898531588845
13 0.0369461512789807
14 0.040090639707164
15 0.0391797806390081
16 0.0410907168215122
17 0.0424871435250117
18 0.0420634920634921
19 0.04386142067674
20 0.0437647575641096
};
\end{axis}

\end{tikzpicture}
    \end{subfigure}
    \hspace{-0.6cm}
    \begin{subfigure}{.02\textwidth}
        \begin{tikzpicture}[scale=0.3]

    \definecolor{custom-green}{RGB}{0,158,115}
    \definecolor{custom-darkblue}{RGB}{0,114,178}
    
    \node[minimum size=2pt, inner sep=0pt] (rap) at (0.5,5.5) {};
    \node[custom-green] at (2.5,5.5) {\tiny RAPNet-PP};
    \node[minimum size=2pt, inner sep=0pt] (gibbs) at (0.5,4.5) {};
    \node[custom-darkblue] at (2.5,4.5) {\tiny Gibbs};
    \node[circle, draw, minimum size=2pt, inner sep=0pt] (circle) at (0.5,3.5) {};
    \node at (2.5,3.5) {\tiny $k=1$};
    \node[rectangle, draw, minimum size=2pt, inner sep=0pt] (rectangle) at (0.5,2.5) {};
    \node at (2.5,2.5) {\tiny $k=2$};
    \node[diamond, draw, minimum size=2pt, inner sep=0pt] (diamond) at (0.5,1.5) {};
    \node at (2.5,1.5) {\tiny $k=3$};
    \node[regular polygon, regular polygon sides=3, draw, minimum size=2pt, inner sep=0pt] (triangle) at (0.5,0.5) {};
    \node at (2.5,0.5) {\tiny $k=4$};

    \draw[custom-green] ($(rap.west)-(5pt,0)$) -- ($(rap.east)+(5pt,0)$);
    \draw[custom-darkblue] ($(gibbs.west)-(5pt,0)$) -- ($(gibbs.east)+(5pt,0)$);
    \draw ($(circle.west)-(6pt,0)$) -- ($(circle.east)+(6pt,0)$);
    \draw[dashed] ($(rectangle.west)-(6pt,0)$) -- ($(rectangle.east)+(6pt,0)$);
    \draw[dash pattern=on 2pt off 2pt] ($(diamond.west)-(6pt,0)$) -- ($(diamond.east)+(6pt,0)$);
    \draw[dotted] ($(triangle.west)-(6pt,0)$) -- ($(triangle.east)+(6pt,0)$);

    \draw (0,0) rectangle (4,6.2);

    \node at (10, -2) {};
\end{tikzpicture}
    \end{subfigure}
    \hspace{1cm}
    \hfill
    \begin{subfigure}{.48\textwidth}
\begin{tikzpicture}

\definecolor{darkgray176}{RGB}{176,176,176}
\definecolor{custom-yellow}{RGB}{230,159,0}
\definecolor{custom-blue}{RGB}{86,180,233}
\definecolor{custom-green}{RGB}{0,158,115}
\definecolor{custom-darkblue}{RGB}{0,114,178}
\definecolor{custom-orange}{RGB}{213,94,0}
\definecolor{custom-mahagony}{RGB}{192,64,0}

\begin{axis}[
tick pos=left,
tick label style={black, font=\tiny},
x grid style={darkgray176},
xmajorgrids,
xmin=0.3, xmax=15.7,
xtick style={color=black},
y grid style={darkgray176},
ylabel={\scriptsize Accuracy},
ylabel style={yshift=-16pt},
ymajorgrids,
ymin=-0.05, ymax=1.05,
ytick style={color=black},
width=1\textwidth,
height=0.56\textwidth,
]

\addplot [thin, custom-green, mark=o, mark size=1.5, mark options={solid,fill opacity=0}]
table {%
1 1
2 0.99836
3 0.99628
4 0.98892
5 0.97284
6 0.94304
7 0.88784
8 0.80888
9 0.71248
10 0.59408
11 0.48048
12 0.3574
13 0.24892
14 0.17284
15 0.10368
};
\addplot [semithick, dashed, custom-green, mark=square, mark size=1.5, mark options={solid,fill opacity=0}]
table {%
1 1
2 0.963345171882329
3 0.956432808345265
4 0.901213660739486
5 0.819836843803291
6 0.71838948384428
7 0.601114613311406
8 0.486787588619832
9 0.380531380945803
10 0.280472722275663
11 0.205022075055188
12 0.135998893193138
13 0.0866995983564932
14 0.0526436255436027
15 0.0265165441176471
};
\addplot [semithick, dash dot, custom-green, mark=diamond, mark size=1.5, mark options={solid,fill opacity=0}]
table {%
1 1
2 0.879768392370572
3 0.836794233739489
4 0.702734223143709
5 0.546641460388728
6 0.407221898660454
7 0.293480130741442
8 0.210303907380608
9 0.152889921580017
10 0.103820068401832
11 0.0694492956106648
12 0.043906965953251
13 0.0255104992190664
14 0.014867522850862
15 0.00576934171810996
};
\addplot [semithick, dotted, custom-green, mark=triangle, mark size=1.5, mark options={solid,fill opacity=0}]
table {%
1 1
2 0.745908028059236
3 0.678555372396224
4 0.476797910807553
5 0.317880216518247
6 0.211275914377212
7 0.135063427576486
8 0.0932743814301505
9 0.0660455830976543
10 0.0416631737781876
11 0.0237431347697507
12 0.0154999578805492
13 0.00858369098712446
14 0.0047066733904858
15 0.00199054491166957
};
\addplot [semithick, custom-darkblue, mark=o, mark size=1.5, mark options={solid,fill opacity=0}]
table {%
1 1
2 1
3 1
4 1
5 1
6 1
7 1
8 1
9 1
10 1
11 1
12 1
13 1
14 1
15 1
};
\addplot [semithick, dashed, custom-darkblue, mark=square, mark size=1.5, mark options={solid,fill opacity=0}]
table {%
1 0.169128814027337
2 0.202104134557395
3 0.221415422376764
4 0.229607677109794
5 0.20878462460248
6 0.174702394631613
7 0.143666346900553
8 0.113801675720468
9 0.0857392825896763
10 0.0707709220832761
11 0.0555555555555556
12 0.0433038184836746
13 0.0341627810350399
14 0.0260013733119707
15 0.0188419117647059
};
\addplot [semithick, dash dot, custom-darkblue, mark=diamond, mark size=1.5, mark options={solid,fill opacity=0}]
table {%
1 0.0272479564032698
2 0.0383174386920981
3 0.0447914879011498
4 0.054554878838927
5 0.0617947657588468
6 0.0640069889341875
7 0.0618728138081312
8 0.0445151953690304
9 0.0388033691548069
10 0.0282882151759318
11 0.0229363138898591
12 0.0176323879125341
13 0.0115693874009371
14 0.00925604535462224
15 0.00721167714763745
};
\addplot [semithick, dotted, custom-darkblue, mark=triangle, mark size=1.5, mark options={solid,fill opacity=0}]
table {%
1 0
2 0.00857365549493375
3 0.0124381837254608
4 0.0184813177983126
5 0.0229614108608346
6 0.0265464351929884
7 0.0255368543238537
8 0.0250829011138509
9 0.0177175178838796
10 0.0169335233464666
11 0.0111533586818758
12 0.00892932356162076
13 0.0063956913237398
14 0.00344595730374853
15 0.00389815045201957
};
\end{axis}
    
\end{tikzpicture}
    \end{subfigure}
    \begin{subfigure}{.48\textwidth}
\begin{tikzpicture}

\definecolor{darkgray176}{RGB}{176,176,176}
\definecolor{custom-green}{RGB}{0,158,115}
\definecolor{custom-darkblue}{RGB}{0,114,178}
\definecolor{custom-orange}{RGB}{213,94,0}
\definecolor{custom-mahagony}{RGB}{192,64,0}

\begin{axis}[
tick pos=left,
tick label style={black, font=\tiny},
x grid style={darkgray176},
xmajorgrids,
xmin=1.1, xmax=20.9,
xtick style={color=black},
xtick distance=2,
y grid style={darkgray176},
ylabel={\scriptsize \textit{wp} score},
ylabel style={yshift=-16pt},
ymajorgrids,
ymin=0, ymax=3.06715165605991,
ytick style={color=black},
width=1\textwidth,
height=0.56\textwidth,
legend style={font=\tiny, at={(0.99,0.9)}
        },
]
\addplot [semithick, custom-green, mark=o, mark size=1.5, mark options={solid,fill opacity=0}]
table {%
2 2.77373901109784
3 2.66807765331849
4 2.61099722860328
5 2.5645986428085
6 2.52046727407219
7 2.48090662626327
8 2.44479541290439
9 2.413105885089
10 2.38312506146865
11 2.35459118044426
12 2.33062073888236
13 2.30336968634528
14 2.28923554777693
15 2.26763909928631
16 2.24902561403761
17 2.23016327029764
18 2.21239892139476
19 2.2001207011583
20 2.18352842030867
};
\addplot [semithick, custom-darkblue, dashed, mark=square, mark size=1.5, mark options={solid,fill opacity=0}]
table {%
2 2.63365787981868
3 2.48257676678896
4 2.4119139785195
5 2.35424061214057
6 2.30766928389109
7 2.26416905443422
8 2.22868578273417
9 2.19639266867459
10 2.16473560946831
11 2.13492322146867
12 2.1144715585658
13 2.08913314481598
14 2.07334851287161
15 2.05692782318162
16 2.04218137550092
17 2.02119584661611
18 2.00798315782829
19 1.9981900498346
20 1.98329336665971
};
\addplot [semithick, custom-orange, dash dot, mark=diamond, mark size=1.5, mark options={solid,fill opacity=0}]
table {%
2 2.62655029164799
3 2.1244444788964
4 2.03585752796464
5 1.97114465501416
6 1.91901011292155
7 1.87885029681263
8 1.84601049552426
9 1.81657228203883
10 1.79154961035497
11 1.76861317182657
12 1.75518230950918
13 1.73597933056341
14 1.72403529695031
15 1.70906329590676
16 1.69774072446799
17 1.68576675515005
18 1.67219824181001
19 1.66690351150575
20 1.65654773246939
};
\addplot [semithick, custom-mahagony, dotted, mark=triangle, mark size=1.5, mark options={solid,fill opacity=0}]
table {%
2 3
3 3
4 3
5 3
6 3
7 3
8 3
9 3
10 3
11 3
12 3
13 3
14 3
15 3
16 3
17 3
18 3
19 3
20 3
};
\end{axis}

\end{tikzpicture}
    \end{subfigure}
    \hspace{-0.6cm}
    \begin{subfigure}{.02\textwidth}
        \begin{tikzpicture}[scale=0.3]

    \definecolor{custom-green}{RGB}{0,158,115}
    \definecolor{custom-darkblue}{RGB}{0,114,178}
    \definecolor{custom-orange}{RGB}{213,94,0}
    \definecolor{custom-mahagony}{RGB}{192,64,0}
    
    \node[circle, draw, minimum size=2pt, inner sep=0pt, custom-green] (circle) at (0.5,3.5) {};
    \node[custom-green] at (2.5,3.5) {\tiny RAPNet-PP};
    \node[rectangle, draw, minimum size=2pt, inner sep=0pt, custom-darkblue] (rectangle) at (0.5,2.5) {};
    \node[custom-darkblue] at (2.5,2.5) {\tiny RAPNet-a};
    \node[diamond, draw, minimum size=2pt, inner sep=0pt, custom-orange] (diamond) at (0.5,1.5) {};
    \node[custom-orange] at (2.5,1.5) {\tiny Gibbs};
    \node[regular polygon, regular polygon sides=3, draw, minimum size=2pt, inner sep=0pt, custom-mahagony] (triangle) at (0.5,0.5) {};
    \node[custom-mahagony] at (2.5,0.5) {\tiny Murty};

    \draw[custom-green] ($(circle.west)-(6pt,0)$) -- ($(circle.east)+(6pt,0)$);
    \draw[dashed, custom-darkblue] ($(rectangle.west)-(6pt,0)$) -- ($(rectangle.east)+(6pt,0)$);
    \draw[dash pattern=on 2pt off 2pt, custom-mahagony] ($(diamond.west)-(6pt,0)$) -- ($(diamond.east)+(6pt,0)$);
    \draw[dotted, custom-orange ] ($(triangle.west)-(6pt,0)$) -- ($(triangle.east)+(6pt,0)$);

    \draw (0,0) rectangle (4,4.2);
    
    \node at (10, -2.5) {};
\end{tikzpicture}
    \end{subfigure}
    \hspace{1.035cm}
    \hfill
    \begin{subfigure}{.48\textwidth}
\begin{tikzpicture}

\definecolor{darkgray176}{RGB}{176,176,176}
\definecolor{custom-green}{RGB}{0,158,115}
\definecolor{custom-darkblue}{RGB}{0,114,178}
\definecolor{custom-orange}{RGB}{213,94,0}
\definecolor{custom-mahagony}{RGB}{192,64,0}

\begin{axis}[
tick pos=left,
tick label style={black, font=\tiny},
x grid style={darkgray176},
xmajorgrids,
xmin=0.3, xmax=15.7,
xtick style={color=black},
y grid style={darkgray176},
ylabel={\scriptsize \textit{wp} score},
ylabel style={yshift=-16pt},
ymajorgrids,
ymin=0, ymax=3.13694407398033,
ytick style={color=black},
width=1\textwidth,
height=0.56\textwidth,
legend style={font=\tiny, at={(0.28,0.57), line width=0.2pt, inner sep=.1pt,}
        },
]
\addplot [semithick, custom-green, mark=o, mark size=1.5, mark options={solid,fill opacity=0}]
table {%
1 3
2 2.95945498596832
3 2.88967551708311
4 2.72100459709935
5 2.52060091460068
6 2.31020337691754
7 2.07666316827063
8 1.83642851603378
9 1.57256189019106
10 1.29289837891195
11 1.02592131519437
12 0.753971619408876
13 0.519309589701593
14 0.353346609352604
15 0.208753220635653
};
\addplot [semithick, custom-darkblue, dashed, mark=square, mark size=1.5, mark options={solid,fill opacity=0}]
table {%
1 2.99967502999455
2 2.91342829202563
3 2.78303537520792
4 2.53197788337052
5 2.25892827000234
6 1.99615991228521
7 1.72168183854911
8 1.4512124575159
9 1.14443770415362
10 0.86876661766801
11 0.618713540469557
12 0.411132282728329
13 0.246545072654039
14 0.146203751170188
15 0.0774498483072966
};
\addplot [semithick, custom-orange, dash dot, mark=diamond, mark size=1.5, mark options={solid,fill opacity=0}]
table {%
1 2.67398835364282
2 2.38653438023448
3 2.09602755256049
4 1.93981561721079
5 1.86404563583769
6 1.82362889146492
7 1.77776324960426
8 1.74492301709268
9 1.70225948219091
10 1.67109911949832
11 1.64998224320341
12 1.62943408474207
13 1.6107653798452
14 1.59111829947364
15 1.57777651359737
};
\addplot [semithick, custom-mahagony, dotted, mark=triangle, mark size=1.5, mark options={solid,fill opacity=0}]
table {%
1 3
2 3
3 3
4 3
5 3
6 3
7 3
8 3
9 3
10 3
11 3
12 3
13 3
14 3
15 3
};
\end{axis}

\end{tikzpicture}
    \end{subfigure}
    \begin{subfigure}{.48\textwidth}
\begin{tikzpicture}

\definecolor{darkgray176}{RGB}{176,176,176}
\definecolor{custom-green}{RGB}{0,158,115}
\definecolor{custom-darkblue}{RGB}{0,114,178}
\definecolor{custom-orange}{RGB}{213,94,0}
\definecolor{custom-mahagony}{RGB}{192,64,0}

\begin{axis}[
tick pos=left,
tick label style={black, font=\tiny},
x grid style={darkgray176},
xlabel={\scriptsize $k_{max}$},
xlabel style={yshift=8.5pt},
xmajorgrids,
xmin=1.1, xmax=20.9,
xtick style={color=black},
xtick distance=2,
y grid style={darkgray176},
ylabel={\scriptsize Cost},
ylabel style={yshift=-16pt},
ymajorgrids,
ymin=-53.2490988566368, ymax=-5.12161776288544,
ytick style={color=black},
width=1\textwidth,
height=0.56\textwidth,
legend style={font=\tiny, at={(0.99,0.9)}
        },
]
\addplot [semithick, custom-green, mark=o, mark size=1.5, mark options={solid,fill opacity=0}]
table {%
2 -8.46564889459619
3 -12.8172693082543
4 -14.3122778108172
5 -15.4480991368349
6 -16.3566289495967
7 -16.9015320004495
8 -17.2330720782219
9 -17.3694677191676
10 -17.7939775075338
11 -17.6018857401294
12 -17.5315231554799
13 -17.3511432430809
14 -17.3271663742568
15 -17.1017509289653
16 -16.8226817562164
17 -16.480843748738
18 -16.1765258940953
19 -15.7814695189792
20 -15.4798799885243
};
\addplot [semithick, custom-darkblue, dashed, mark=square, mark size=1.5, mark options={solid,fill opacity=0}]
table {%
2 -7.76334021055621
3 -11.0284360766309
4 -11.6935087529696
5 -11.982278069321
6 -12.0729901329828
7 -11.9255635620699
8 -11.6823784554156
9 -11.3338343585608
10 -11.1358051397369
11 -10.6731917290685
12 -10.2829261354936
13 -9.818831930066
14 -9.53366556602938
15 -9.18350178470439
16 -8.76383967755811
17 -8.23859627009162
18 -7.7664916736035
19 -7.33541713635585
20 -6.90241119119654
};
\addplot [semithick, custom-orange, dash dot, mark=diamond, mark size=1.5, mark options={solid,fill opacity=0}]
table {%
2 -7.30466240913761
3 -9.47713639943094
4 -10.5325046251024
5 -11.5118447930853
6 -12.5735390377119
7 -13.478176269858
8 -14.3322785401539
9 -15.1836690701971
10 -16.1667831251636
11 -17.0366880090233
12 -17.8524801609182
13 -18.7623813301704
14 -19.4704687672889
15 -20.4889216186361
16 -21.0601413179655
17 -21.8752344176548
18 -22.6527473940271
19 -23.3958623230535
20 -23.9714708415141
};
\addplot [semithick, custom-mahagony, dotted, mark=triangle, mark size=1.5, mark options={solid,fill opacity=0}]
table {%
2 -9.06988439871343
3 -14.4826453176735
4 -17.010981383556
5 -19.4028111172217
6 -21.8927144774174
7 -24.1466317616183
8 -26.3491074142567
9 -28.5019958280872
10 -30.8098564301384
11 -33.0403384890383
12 -35.0838170225056
13 -37.2325107455953
14 -39.027019573999
15 -41.3899171132558
16 -43.0458254921719
17 -45.1200588481732
18 -47.1447271093058
19 -49.1711615313534
20 -51.061486079648
};
\end{axis}

\end{tikzpicture}
    \end{subfigure}
    \hspace{-0.6cm}
    \begin{subfigure}{.02\textwidth}
        \begin{tikzpicture}[scale=0.3]

    \definecolor{custom-green}{RGB}{0,158,115}
    \definecolor{custom-darkblue}{RGB}{0,114,178}
    \definecolor{custom-orange}{RGB}{213,94,0}
    \definecolor{custom-mahagony}{RGB}{192,64,0}
    
    \node[circle, draw, minimum size=2pt, inner sep=0pt, custom-green] (circle) at (0.5,3.5) {};
    \node[custom-green] at (2.5,3.5) {\tiny RAPNet-PP};
    \node[rectangle, draw, minimum size=2pt, inner sep=0pt, custom-darkblue] (rectangle) at (0.5,2.5) {};
    \node[custom-darkblue] at (2.5,2.5) {\tiny RAPNet-a};
    \node[diamond, draw, minimum size=2pt, inner sep=0pt, custom-orange] (diamond) at (0.5,1.5) {};
    \node[custom-orange] at (2.5,1.5) {\tiny Gibbs};
    \node[regular polygon, regular polygon sides=3, draw, minimum size=2pt, inner sep=0pt, custom-mahagony] (triangle) at (0.5,0.5) {};
    \node[custom-mahagony] at (2.5,0.5) {\tiny Murty};

    \draw[custom-green] ($(circle.west)-(6pt,0)$) -- ($(circle.east)+(6pt,0)$);
    \draw[dashed, custom-darkblue] ($(rectangle.west)-(6pt,0)$) -- ($(rectangle.east)+(6pt,0)$);
    \draw[dash pattern=on 2pt off 2pt, custom-mahagony] ($(diamond.west)-(6pt,0)$) -- ($(diamond.east)+(6pt,0)$);
    \draw[dotted, custom-orange ] ($(triangle.west)-(6pt,0)$) -- ($(triangle.east)+(6pt,0)$);

    \draw (0,0) rectangle (4,4.2);
    
    \node at (10, -3.6) {};
\end{tikzpicture}
    \end{subfigure}
    \hspace{1.035cm}
    \hfill
    \begin{subfigure}{.48\textwidth}
\begin{tikzpicture}

\definecolor{darkgray176}{RGB}{176,176,176}
\definecolor{custom-green}{RGB}{0,158,115}
\definecolor{custom-darkblue}{RGB}{0,114,178}
\definecolor{custom-orange}{RGB}{213,94,0}
\definecolor{custom-mahagony}{RGB}{192,64,0}

\begin{axis}[
tick pos=left,
tick label style={black, font=\tiny},
x grid style={darkgray176},
xlabel={\footnotesize $\nu_s$},
xlabel style={yshift=7pt},
xmajorgrids,
xmin=0.3, xmax=15.7,
xtick style={color=black},
y grid style={darkgray176},
ylabel={\scriptsize Cost},
ylabel style={yshift=-16pt},
ymajorgrids,
ymin=-127.092335980529, ymax=15,
ytick style={color=black},
width=1\textwidth,
height=0.56\textwidth,
legend style={font=\tiny, at={(0.28,0.57)}
        },
]
\addplot [semithick, custom-green, mark=o, mark size=1.5, mark options={solid,fill opacity=0}]
table {%
1 -0.47718678599511
2 -2.18268474861572
3 -6.43686696971963
4 -12.0474153140437
5 -18.3384093275383
6 -24.4646806001961
7 -29.5002450397928
8 -33.2692281273674
9 -36.1332690320856
10 -36.9026295686442
11 -35.7065919012727
12 -30.9095274968786
13 -23.6293189141335
14 -15.2752983340503
15 -5.13058757273838
};
\addplot [semithick, custom-darkblue, dashed, mark=square, mark size=1.5, mark options={solid,fill opacity=0}]
table {%
1 -0.476360900911823
2 -2.03698457413535
3 -5.94014371752698
4 -10.6625508881306
5 -15.5636840094279
6 -19.5200847835558
7 -22.0180308910258
8 -22.694295059873
9 -21.6267324331345
10 -18.5213087953181
11 -13.6141986025549
12 -7.10320332448346
13 -0.355709618748628
14 5.5216870671538
15 10.8448220987733
};
\addplot [semithick, custom-orange, dash dot, mark=diamond, mark size=1.5, mark options={solid,fill opacity=0}]
table {%
1 -0.367099835807214
2 -1.18271693030309
3 -3.14448830633457
4 -6.13966863357377
5 -10.6322663984805
6 -16.8101298030855
7 -24.0246031150088
8 -32.1242527333041
9 -41.1242327783058
10 -50.1117456968537
11 -59.4046659863765
12 -69.4399874130742
13 -78.7453304168611
14 -88.7695847898027
15 -99.1898173274306
};
\addplot [semithick, custom-mahagony, dotted, mark=triangle, mark size=1.5, mark options={solid,fill opacity=0}]
table {%
1 -0.47718678599511
2 -2.28232957365191
3 -6.8324220537066
4 -13.4518561883541
5 -21.6723422382073
6 -30.8369851355384
7 -40.2692187524189
8 -49.6005197304293
9 -59.6003306874013
10 -69.3378880561122
11 -79.0304407597106
12 -89.3876862311193
13 -99.2694173518656
14 -109.87491944917
15 -120.780701869678
};
\end{axis}

\end{tikzpicture}
    \end{subfigure}
    \vspace{-20pt}
    \caption{Plots of the two sweeps of the parameters $k_{\text{max}}$ and $\nu_s$ evaluated with the scores accuracy, \textit{wp} and cost. For readability reasons in the accuracy plots, only the accuracies of the first $4$ solutions for the RAPNet with the post-processing and the Gibbs sampler are shown. The legends on the right are also valid for the left plots. In the legends, RAPNet-PP and RAPNet-a indicate the versions with and without post-processing, respectively.}
    \label{sweeps}
\end{figure*}
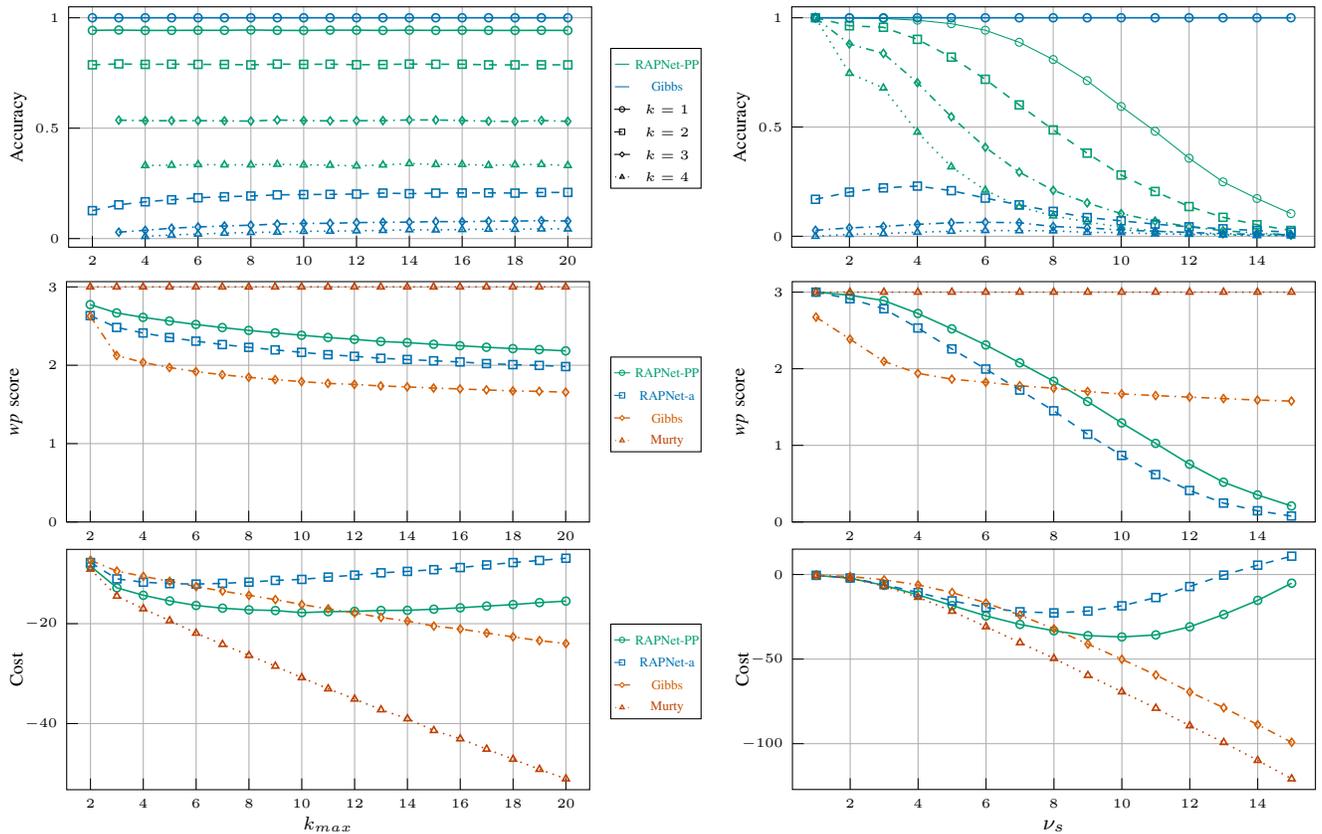

The plots of the described sweeps are shown in Fig.~\ref{sweeps}.

Note that the accuracy score is only computable for assignment $i$ if $k\geq i$. Also, the accuracy plots of Murty's algorithm and the RAPNet-a are excluded for readability reasons.

The plots on the left show the influence of the maximum number of $k$, i.e., $k_{\text{max}}$.
On the left side, the accuracy plot demonstrates that the accuracies of the first $4$ assignments are practically independent of $k_{\text{max}}$.
This applies to both the Gibbs sampler and the RAPNet-PP, with the RAPNet-PP being generally more accurate than the Gibbs sampler except for $k=1$, where the Gibbs sampler always finds the optimal solution that is used as the initial assignment.
Concerning the \textit{wp} scores on the left, both the RAPNet-PP and the RAPNet-a outperform the Gibbs sampler.
The decreasing \textit{wp} score for higher values of $k_{\text{max}}$ for both the RAPNet and the Gibbs sampler can be explained by the fact that the weight value is smaller for the first assignments with a higher $k_{\text{max}}$, but the accuracy of both methods decreases for assignments of higher $k$.
The average costs of the $k$ assignments can be seen in the plot at the bottom left.
Both the \textit{wp} score and the cost plot show that the developed post-processing algorithm is able to improve the predictions of the RAPNet-a.
The costs are often negative, which follows from the derived Gaussian Mixture model.
For lower $k_{\text{max}}$, the two RAPNet versions perform better than the Gibbs sampler.
The Gibbs sampler surpasses the RAPNet-PP for $k_{\text{max}}>11$ and the RAPNet-a for $k_{\text{max}}>5$.
One reason why the RAPNet is worse for higher values of $k_{\text{max}}$ is obviously the design choice of using a fixed number of assignments to be predicted.
Thus, the cost of the RAPNet-a has to increase for $k>k_{\text{max}}$, while the RAPNet-PP potentially creates more than just $k=10$ solutions.
In combination with the other plots, it can be concluded that the Gibbs sampler finds more solutions for higher $k$ which, however, are usually not the optimal solutions.
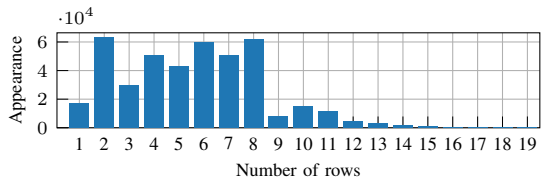
\begin{figure}[t]
    \centering
    \vspace{-0.4cm}
\begin{tikzpicture}

\definecolor{darkgray176}{RGB}{176,176,176}
\definecolor{steelblue31119180}{RGB}{31,119,180}

\begin{axis}[
tick pos=left,
tick label style={black, font=\scriptsize},
x grid style={darkgray176},
xlabel={\scriptsize Number of rows},
xlabel style={yshift=5pt},
xmajorgrids,
xmin=-0.9, xmax=18.5,
xtick style={color=black},
xtick={0,1,2,3,4,5,6,7,8,9,10,11,12,13,14,15,16,17,18},
xticklabels={1,2,3,4,5,6,7,8,9,10,11,12,13,14,15,16,17,18,19},
y grid style={darkgray176},
ylabel={\scriptsize Appearance},
ylabel style={yshift=-20pt},
ymajorgrids,
ymin=0, ymax=66416.7,
ytick style={color=black},
width=0.45\textwidth,
height=0.16\textwidth,
]
\draw[draw=none,fill=steelblue31119180] (axis cs:-0.4,0) rectangle (axis cs:0.4,17204);
\draw[draw=none,fill=steelblue31119180] (axis cs:0.6,0) rectangle (axis cs:1.4,63254);
\draw[draw=none,fill=steelblue31119180] (axis cs:1.6,0) rectangle (axis cs:2.4,29586);
\draw[draw=none,fill=steelblue31119180] (axis cs:2.6,0) rectangle (axis cs:3.4,50382);
\draw[draw=none,fill=steelblue31119180] (axis cs:3.6,0) rectangle (axis cs:4.4,43206);
\draw[draw=none,fill=steelblue31119180] (axis cs:4.6,0) rectangle (axis cs:5.4,59841);
\draw[draw=none,fill=steelblue31119180] (axis cs:5.6,0) rectangle (axis cs:6.4,50650);
\draw[draw=none,fill=steelblue31119180] (axis cs:6.6,0) rectangle (axis cs:7.4,61620);
\draw[draw=none,fill=steelblue31119180] (axis cs:7.6,0) rectangle (axis cs:8.4,8304);
\draw[draw=none,fill=steelblue31119180] (axis cs:8.6,0) rectangle (axis cs:9.4,14776);
\draw[draw=none,fill=steelblue31119180] (axis cs:9.6,0) rectangle (axis cs:10.4,11214);
\draw[draw=none,fill=steelblue31119180] (axis cs:10.6,0) rectangle (axis cs:11.4,4452);
\draw[draw=none,fill=steelblue31119180] (axis cs:11.6,0) rectangle (axis cs:12.4,3204);
\draw[draw=none,fill=steelblue31119180] (axis cs:12.6,0) rectangle (axis cs:13.4,1812);
\draw[draw=none,fill=steelblue31119180] (axis cs:13.6,0) rectangle (axis cs:14.4,0954);
\draw[draw=none,fill=steelblue31119180] (axis cs:14.6,0) rectangle (axis cs:15.4,0506);
\draw[draw=none,fill=steelblue31119180] (axis cs:15.6,0) rectangle (axis cs:16.4,0232);
\draw[draw=none,fill=steelblue31119180] (axis cs:16.6,0) rectangle (axis cs:17.4,0104);
\draw[draw=none,fill=steelblue31119180] (axis cs:17.6,0) rectangle (axis cs:18.4,0020);
\end{axis}

\end{tikzpicture}
    \vspace{-10pt}
    \caption{Distribution of matrix sizes in the simulation data.}
    \label{fig:sim_sizes}
    \vspace{-0.55cm}
\end{figure}

All three of the plots of the size sweep of Fig.~\ref{sweeps} on the right show a performance decrease of both the RAPNet and the Gibbs sampler w.r.t.\ bigger matrices.
Compared with the Gibbs sampler, both RAPNet versions have better \textit{wp} scores and lower cost values for $\nu_s<9$ or $\nu_s<8$, respectively.
The Gibbs sampler outperforms the RAPNet versions for bigger matrices.
This can be explained with the fast decline of the accuracy curves of the RAPNet seen in the accuracy plot on the right, while the Gibbs sampler especially profits from using the optimal solution for $k=1$.
However, for the considered MOT scenario, matrices with more than $8$ rows are very rare.
This can clearly be seen in Fig. \ref{fig:sim_sizes}, where the matrix appearance frequency in relation to the number of rows of the simulation dataset is plotted.

Thus, a comparison of the framework's performance to the Gibbs sampler on a validation set that only includes simulation data is presented in Table \ref{table:simulation_eval}. 
While the Gibbs sampler generates the optimal solution for the accuracy of the initial assignment, both RAPNet versions perform better in all other scores.
The post-processing improves the already good performance of the RAPNet itself.
With Murty's algorithm, the optimal accuracy scores, \textit{wp} score and cost value are $1.0$, $3.0$ and $-5.92$, respectively.

\begin{table}
    \centering
    \vspace{-10pt}
    \caption{Comparison of the RAPNet versions to the Gibbs sampler on simulation data.}
    \begin{tabular}{@{}l | c c c c c c@{}}
        \multirow{2}{*}{Framework} & \multicolumn{4}{c}{Accuracies} & \multirow{2}{*}{\textit{wp}} &\multirow{2}{*}{Cost} \\
         & {\scriptsize $i=1$} & {\scriptsize $i=2$} & {\scriptsize $i=3$} & {\scriptsize $i=4$} & & \\
         \hline
         RAPNet-a & $\underline{0.99}$ & $\underline{0.91}$ & $\underline{0.73}$ & $\underline{0.54}$ & $\underline{2.74}$ & $\underline{5.68}$ \\
         RAPNet-PP & $\underline{0.99}$ & $\mathbf{0.95}$ & $\mathbf{0.82}$ & $\mathbf{0.66}$ & $\mathbf{2.80}$ & $\mathbf{3.23}$ \\
         Gibbs & $\mathbf{1}$ & $0.18$ & $0.06$ & $0.04$ & $2.11$ & $14.10$ \\
    \end{tabular}
    \label{table:simulation_eval}
    \vspace{-0.55cm}
\end{table}

\subsection{Computational Complexity}
Lastly, the needed computational time of the parts of the framework, the Gibbs sampler and Murty's algorithm w.r.t.\ the matrix sizes is summarized in Table \ref{table:modulecomp}.
For the RAPNet and the post-processing, the computational time is taken for non-batched and batched data with a batch size of $32$, i.e., RAPNet$_{32}$ and PP$_{32}$.
The experiments were done on an AMD Ryzen 9 7950x CPU for Murty's algorithm and the Gibbs sampler and an NVIDIA RTX 4080 GPU for the parts of the RAPNet framework.
The comparison shows that, without batching, the RAPNet's needed computational time is generally higher compared to the Gibbs sampler and Murty's algorithm.
For the batched version, except for very small matrices with $\nu_s<3$, the computational time of RAPNet$_{32}$ for each single graph in the batch is similar to that of the Gibbs sampler.
The table also shows the increased slowdown of Murty's algorithm for bigger matrices.
For the MOT application, the overhead of the graph creation step can be reduced by directly creating graphs instead of cost matrices.
For the unbatched version, the post-processing function adds a relatively high overhead that is less for the batched version.
We argue that the worse computational overhead for unbatched graphs results from the complexity of the GNN layers.
This fact is supported by an investigation into the ratio of the needed time of the RAPNet's encoder to the decoder, which shows that the encoder takes about $4/5$th of the RAPNet's computational time.
However, the practically constant time of the RAPNet for unbatched graphs shows an advantage compared to the other methods, where the needed time increases for bigger graphs.
It is part of our future work to resolve the current limitations concerning the computational complexity of the framework.

\begin{table}
\vspace{7pt}
\centering
\caption{Time consumption of the different modules, in ms.}
\begin{tabular}{@{}c | c | c | c | c | c | c | c@{}}
    Rows &
      Graph & RAP- & RAP- & \multirow{2}{*}{PP$_{32}$} & \multirow{2}{*}{PP} & \multirow{2}{*}{Gibbs} & \multirow{2}{*}{Murty} \\
    $\nu_s$ & Creation & Net$_{32}$& Net & & & & \\
    \hline
    $1$ & 0.29 & 0.12 & 3.04 & 0.32 & 0.53 & 0.05 & 0.03 \\
    $3$ & 0.30 & 0.15 & 3.03 & 0.39 & 0.59 & 0.12 & 0.14 \\
    $5$ & 0.30 & 0.19 & 3.06 & 0.45 & 0.64 & 0.19 & 0.28 \\
    $7$ & 0.31 & 0.25 & 3.11 & 0.47 & 0.68 & 0.26 & 0.43 \\
    $9$ & 0.31 & 0.32 & 3.15 & 0.48 & 0.70 & 0.32 & 0.61 \\
    $11$ & 0.31 & 0.39 & 3.15 & 0.48 & 0.71 & 0.38 & 0.83 \\
    $13$ & 0.30 & 0.48.11 & 0.48 & 0.69 & 0.43 & 1.06 \\
    $15$ & 0.33 & 0.58 & 3.22 & 0.48 & 0.69 & 0.50 & 1.38 \\
\end{tabular}
\label{table:modulecomp}
\vspace{-0.55cm}
\end{table}

\section{Conclusion and Future Work} \label{sec:Conclusion}
This paper presented a novel approach utilizing a Graph Neural Network for solving the ranked assignment problem, with a particular focus on assignment problems within the update step of the $\delta$-GLMB filter.
The proposed RAPNet was compared to the Gibbs sampler and Murty's algorithm on both data from a simulated MOT scenario and synthetic data, showing the capabilities of the RAPNet itself and in combination with the post-processing stage, but also some of the current restrictions.

Our work can be the foundation of further research in this field.
In our own future work, we want to reduce the computational complexity and integrate our approach into our tracking framework.
As motivated in the introduction, we will also extend the framework to handle assignment problems that result from MS-MOT scenarios.
For MS-MOT scenarios, it is also very interesting how the computational complexities of the different algorithms scale, especially due to the NP-hard complexity of getting an optimal solution.
An investigation into using not only the cost values as inputs to the network, but also explicit track representations, could also be worthwhile. Here, due to its flexibility, our GNN-based approach could make use of the additional features that are not used by the Gibbs sampler.


\bibliographystyle{IEEEtran}
{\footnotesize
	\bibliography{RAPNet}}

\end{document}